\documentclass[11pt]{article}

\usepackage[preprint]{acl}

\usepackage{times}
\usepackage{latexsym}
\usepackage[T1]{fontenc}
\usepackage[utf8]{inputenc}
\usepackage{microtype}
\usepackage{inconsolata}
\usepackage{graphicx}
\usepackage{booktabs}
\usepackage{tabularx}
\usepackage{amsmath}
\usepackage{amsfonts}
\usepackage{nicefrac}
\usepackage{url}
\usepackage{xurl}

\AtBeginEnvironment{thebibliography}{\emergencystretch=1em}

\title{Toxicity Is a Property of Communicative Events: A Contextual Stress Measurement Framework for NLP}

\author{Sergei Berezin \and Reza Farahbakhsh \and Noel Crespi \\
  SAMOVAR, Télécom SudParis, Institut Polytechnique de Paris \\
  Palaiseau, France \\
  \texttt{sergeyberezin123@gmail.com}}

\begin{document}

\maketitle

\begin{abstract}
Toxicity detection has become core safety infrastructure for online moderation, dataset filtering, and deployed language-model systems. Yet most detectors still treat toxicity as an intrinsic property of isolated text. \textbf{This position paper argues that toxicity detection should be evaluated as the contextual measurement of situated communicative harm, rather than as single-label text classification.} Toxicity is not contained in words alone; it emerges when a communicative act is interpreted by an audience within a normative and social context.

We introduce the Contextual Stress Framework (CSF), which defines toxicity as a relation between perceived norm violation and induced stress or disruption. CSF explains why text-intrinsic detectors overflag dialectal or reclaimed language, miss coded or pragmatic abuse, and remain brittle under meaning-preserving transformations.
\end{abstract}

\section{Introduction}

Toxicity detection has become a critical component of modern online communication. It operates inside content-moderation pipelines, helping platforms assess harm, filter comments, and decide which content is removed, down-ranked, or escalated to human reviewers. Increasingly, it is also embedded within language models' pipelines: as data filters during pretraining and as guard models in deployment. Toxicity detection is therefore no longer a niche NLP task; it now functions as operational interaction infrastructure.

When a system becomes infrastructure, its failures carry greater consequences. A brittle detector is not merely "a model with lower F1" - it is a weak link in systems intended to protect individuals in vulnerable settings. If a system fails to detect genuinely harmful content (false negatives), it can enable escalation, coordinated abuse, and harassment. If it over-predicts toxicity (false positives), it will penalise benign speech, misinterpret dialects or reclaimed language, and discourage participation. Both types of mistakes impose distinct and significant social costs.

Systems that limit harassment simultaneously shape the boundaries of acceptable discourse. Automated moderation can contribute to echo chambers or amplify dominant perspectives. Toxicity detection systems operate as socio-technical control mechanisms. When they implicitly encode cultural bias, political assumptions, or majority norms, they become instruments of exclusion.

With the importance being this high, the more surprising is to see that many toxicity systems are built on a narrow abstraction: toxicity is treated as an intrinsic property of isolated text. This abstraction is convenient for annotation, benchmarking, and deployment, but it collapses the conditions that determine whether communication is harmful: who speaks, who is addressed, when and where the exchange occurs, what prior interaction exists, which norms apply, and how the message is received.

As a result, benchmark progress can be a misleading indicator of safety. Higher accuracy on decontextualised labels may simply mean that a model has learned dataset-specific conventions: which words, topics, identities, sentiments, or stylistic patterns annotators associated with toxicity. But moderation systems are not deployed to reproduce annotation conventions. They are deployed to reduce harm.

Therefore, \textbf{this position paper argues that toxicity detection should be designed as robust, context-aware, governable safety infrastructure for human communication – judged not just by fit to static labels, but by alignment with real harm under real context and variation.}

Toxicity should be operationalised not as a property contained in text alone, but as a contextual and socially emergent relation between a communicative act, an interpreting audience, applicable norms, and experienced or observable stress.

We introduce the \emph{Contextual Stress Framework} (CSF), which frames toxicity as a stress-inducing norm violation \emph{within a given context}. CSF changes the object of evaluation from text classification to evaluation of a communicative event.

This paper makes four contributions. First, it distinguishes context-aware prediction from context-constituted measurement, arguing that context is part of the toxicity construct rather than merely an additional feature of an input. Second, it formalises toxicity at the level of audience-specific communicative events through perceived norm violation and attributable stress. Third, it separates individual contribution, audience aggregation, and institutional enforcement. Fourth, it derives contrastive evaluation criteria for toxicity detection systems.

\section{The Object Problem and the Proxy Problem}

Current toxicity detection faces two related problems. First, it lacks a settled object: legal systems, platforms, datasets, and research communities use overlapping but non-equivalent notions of toxic, hateful, abusive, offensive, uncivil, and harmful speech. Second, even when a definition is chosen, it is usually operationalised as an utterance-to-label mapping. Toxicity becomes a function of an isolated string rather than a measurement of situated communicative harm.

\subsection{The target is unclear}

When asked to define obscenity, Justice Potter Stewart famously said: "I know it when I see it"~\citep{jacobellis1964}.
To this day, we have not come far beyond this intuition.

Natural language processing (NLP) has reported remarkable technical progress in detecting "toxic" language. Yet, despite ever-larger models and rising benchmark scores, a foundational question remains unresolved: \textbf{what exactly is toxicity?}

An influential work by \citet{def_old} defines toxic speech as "rude, disrespectful, or unreasonable language likely to make someone leave a discussion". Terms such as "rude" and "disrespectful" are inherently subjective: they vary across individuals, communities, cultures, and interactional settings. As a result, this causes annotators to disagree even on simple phrases such as "I don't like you", which some may label as toxic and others as not~\citep{authors1,autors2,authors3}. More than that, such definitions offer no quantitative basis for measurement.

At the level of academia as a whole, the lack of a standardised working definition has led researchers to employ different terms for similar phenomena \cite{vidgen-etal-2019-challenges}, assigning labels such as "toxic", "hateful", "offensive", or "abusive" to effectively the same data \cite{fortuna2020toxic_hateful_offensive,madukwe-etal-2020-data}. In this paper, we therefore use \emph{toxic speech} as an umbrella term.

Stakeholders, such as legal systems, platforms, and academia, have each developed their own definitions of harmful online speech~\citep{brown2017hate_speech_systematic,mnassri2024multilingual_offensive_survey,kiritchenko2021ethical_human_rights,coe1997r9720,coe2022cmrec16,un2019strategy_hate_speech}. As a result, the same utterance can be legally protected, removable under platform policy, and labelled both "toxic" and "non-toxic" in different datasets without any actor being inconsistent. A detailed comparison of these definitional approaches is provided in Appendix~\ref{appA}.

The field therefore lacks a stable object of measurement. 

\subsection{Operational mismatch}
\label{subsec:operational-mismatch}

Even with definition being fixed, toxicity is commonly operationalised as
\(
y = f(x),
\)
where \(x\) is an isolated text and \(y\) is a toxicity label or score. This formulation is computationally convenient, but it collapses audience, norms, setting, relationship, prior interaction, and reception into a single decontextualised prediction.

This mismatch explains several recurring failures. Text-only detectors are prone to racial, sexual, political, religious, and geographical biases~\citep{survey,bias_nig}. They behave like profanity filters~\citep{holy_shit}, lack nuance in real-world use~\citep{xenos-etal-2021-context,context2,context3}, and remain vulnerable to adversarial transformations~\citep{berezin-etal-2025-tip,berezin2024,berezin-etal-2023-offence}. 

What this approach failles to capture, is that in real life the same string can function as abuse, affiliation, quotation, reclamation, ordinary reference, or a non-offensive lexical item depending on who says it, to whom, where, and when. A slur may be an out-group insult, an in-group solidarity marker, a reclaimed identity term, a quotation in a moderation appeal, or an object of academic discussion~\citep{rahman2012nword,smith2019nigga,galinsky2013reappropriation,kiritchenko2021ethical_human_rights}. 

The problem becomes even more obvious across languages and cultures. A lot of toxicity detectors are multilingual in a sense that they can process multiple languages and scripts, yet they remain monolingual and monocultural in their assumptions. In some cultures and cases, profanity can mark solidarity or humour; in others, mild-looking disrespect toward elders, teachers, religious figures, or superiors can be more disruptive than explicit profanity~\citep{brown_levinson1987,dewaele2004}. Honorific systems, religious taboos, caste-marked language, and region-specific norms show that norm violation is often encoded in context, register, relation, or social meaning rather than in "universally bad words" ~\citep{ide1989discernment,brown2011korean_honorifics,brown_gilman1960,temperman_koltay2017_blasphemy,coe2022cmrec16,un2019strategy_hate_speech}.

Thus, just assigning a label to an isolated text, be this "toxic", "offensive", "abusive", "hateful", or "uncivil", - does not solve the measurement problem. A new framework is needed to specify what must be measured when a communicative act is judged toxic.

\subsection{Context-aware classification is not enough}
\label{sec:context-aware-not-enough}

A natural response is to make toxicity detection more context-aware. This is an important direction: prior work shows that conversational context affects both annotation and prediction in abusive-language, hate-speech, and toxicity detection~\citep{gao2017detecting,vidgen2021cad,context2,context3}.

Concurrent to our work, \citet{anonymous_reception_2026} empirically evaluate negative replies as a concrete community-reception signal and compare it with text-intrinsic toxicity scores. In contrast, we propose a theoretical framework that conceptualises toxicity through perceived norm violation and attributable audience stress.

However, context-aware classification does not by itself solve the object problem. Existing approaches incorporate different forms of context: parent comments, conversation history, annotator rationales, speaker information, target identity, platform setting, or community background - but often treat these as auxiliary features rather than as part of the construct being measured. The underlying abstraction remains:
\(
y = f(x, C),
\)
where \(x\) is text, \(C\) is context, and \(y\) is still a toxicity label or score.

This improves prediction, but it does not change the target. If toxicity is treated as a fixed property to be recovered from text, then context just used to infer that property more accurately.

\section{Towards a New Foundation: Cross-Field View on Toxic Speech}
\label{sec:ch7_multifield_insights}

Once stated plainly, the real objective of toxicity detection is not hard to identify. Moderation systems are not deployed to detect profanity, to reproduce annotator averages, or to maximise benchmark F1 scores. They are deployed \textbf{to reduce harm}.

What matters is not whether a text contains a suspicious word pattern, but whether a communicative act contributes to the kinds of social conditions moderation is intended to prevent. These include chronic stress, perceived unsafety, humiliation, escalation, silencing, and the erosion of conditions for participation. At the community level, the long-term objective is healthier discourse: fewer attacks that drive people away, fewer false alarms on normal in-group communication, and closer alignment between what systems flag and what communities actually experience as harmful.

An interaction-aware theory of toxicity does not need to be built from scratch. Across psychology, neuroscience, and sociolinguistics, verbal harm is consistently understood as fundamentally situational: the same utterance can be harmless or harmful depending on who interprets it, under what expectations, and within which normative setting.

\subsection{Individual level}

The experience of verbal harm is fundamentally affective, rooted in how individuals perceive threats to personal and social stability. Research in psychology and neuroscience consistently shows that such harm does not arise from the intrinsic properties of language alone, but from context-dependent appraisals, emotional expectations, and violations of predictive frameworks.

\textbf{In psychology}, appraisal theory \citep{lazarus1991emotion} posits that emotional and stress responses arise when individuals evaluate events as threatening to their goals, values, or social standing. From this perspective, communication becomes harmful not solely because of its content, but because it is perceived as violating significant personal concerns.

Research on social pain further demonstrates that symbolic threats can activate biological stress systems in ways similar to physical injury \citep{eisenberger2003does,lieberman2009neuroscience}.

\textbf{Neuroscientific findings} indicate that the brain processes language in an anticipatory, predictive, and context-sensitive manner. Predictive processing models \citep{friston2010free,kleckner2017evidence,barrett2015interoceptive} suggest that the brain continuously generates expectations about emotional and social states. When communication violates these expectations, it can trigger stress responses.

Physiological indicators of stress, such as heart-rate variability, are influenced by exposure to social threat \citep{shaffer2014healthy}. Event-related potential studies \citep{ding2016emotion,liang2018contextual} show that negatively valenced language can disrupt neural integration, while chronic exposure to social stressors has been associated with pro-inflammatory genetic activity \citep{slavich2013emerging}.

Together, this body of research demonstrates that responses to harmful or toxic speech are not merely psychological but also physiological.

\subsection{Group level}

Beyond direct victimisation, witnessing harm inflicted on others can also elicit stress responses, signalling potential threats within the social environment \citep{struiksma2022insulting}.

\textbf{Sociology and social psychology} show that responses to verbal behaviour are governed by group norms, emotional expectations, and mechanisms of norm enforcement \citep{elster2007explaining}. Research by \citet{Bicchieri_2005} and \citet{CIALDINI1991201} demonstrates that social norms are maintained through collective emotional and behavioural reactions to perceived deviance. Importantly, norm violations are rarely evaluated in isolation; they are interpreted through the emotional dynamics of the group, which shape perceptions of harm.

The concepts of emotional collectives \citep{vankleef2015emotions} and emotional norm-shaping processes \citep{VANKLEEF2024,sari2024role} further emphasise how shared emotional responses reinforce or reshape social expectations in real time.

\textbf{Moral Foundations Theory} likewise suggests that emotional reactions to norm violations differ across groups and cultures, leading to divergent judgements about what constitutes harmful or toxic speech \citep{GRAHAM201355}.

From this perspective, toxicity functions as a social signal: a disruption in the moral and emotional fabric of the group.

\subsection{Cultural and linguistic level}

Linguistic and sociolinguistic research demonstrates that meaning is co-constructed in interaction and shaped by contextual framing. Harmful speech cannot be reduced to individual words or expressions detached from their social use.

\textbf{Pragmatics} \citep{clark1996using,Sperber1995-SPER} emphasises that meaning is not fully encoded in words but inferred through shared background knowledge and conversational context. An identical phrase may be benign in one setting and deeply offensive in another, depending on tone, power relations, and prior discourse.

\textbf{Politeness and impoliteness theory} \citep{Brown_Levinson_1987,CULPEPER1996349} similarly shows that perceived rudeness, offence, or aggression are emergent properties of interaction rather than intrinsic features of utterances.

\textbf{Language ideologies} \citep{woolard1994language,irvine2001style} further demonstrate that judgements about what counts as "toxic" or "neutral" language are shaped by power relations and dominant cultural narratives.

\textbf{Historical linguistics and semantics} highlight the instability of meaning across time and context \citep{allan2006forbidden,Traugott_Dasher_2001}. Words such as "nice" (originally meaning "foolish") or "fag" (benign in British English but a slur in American usage) illustrate how emotional and moral associations shift across cultural and historical settings, undermining any fixed universal standard of toxicity.

Finally, \textbf{multilingualism and language variation} introduce further complexity. Emotional resonance and perceived offensiveness vary across languages due to differing patterns of socialisation \citep{Dewaele01062004}. What is taboo or offensive in one language may carry no such connotation in another.

Taken together, these perspectives argue strongly against treating harmful language as a static or text-intrinsic phenomenon. Toxicity cannot be reliably detected by identifying "bad words" in isolation, {but only by modelling how a communicative act is interpreted within a specific social setting.}




\section{Core intuition of the Contextual Stress Framework}
\label{sec:ch8_core_intuition}

The Contextual Stress Framework is based on a simple postulate: a communicative act becomes toxic when, in a given context, it is \emph{perceived} as violating relevant moral or interactional norms and this perception induces stress in an audience.







Consider the following example: a speaker writes a profane, intentionally offensive message in Chinese to a recipient who does not speak the language. Although the intent is hostile and the string contains profanity, no harm is experienced if the recipient cannot interpret it. There is no perceived norm violation, no affective response, and no stress. Under our framework, such a message registers as low in toxicity in this interaction - not because it is morally defensible, but because it fails to produce a measurable signal of harm in the audience that actually receives it. The same message would become toxic if it reached an informed audience.

Conversely, imagine a speaker makes a flirtatious joke believing it to be benign or even complimentary. If the recipient experiences it as inappropriate or violating - due to context, power dynamics, or personal boundaries - stress may occur. In many professional and legal contexts, this is recognised as harm: offence is taken, not given. Under our model, such an interaction can be toxic even in the absence of harmful intent.

These examples motivate a crucial point: toxicity is defined not by intent or language alone, but by its reception - by the stress it induces and the norms it is perceived to violate.

\section{Definitions}
\label{sec:ch8_definitions}

On this basis, we propose the following definitions.

\begin{quote}
\textbf{Definition 1 (Toxicity).} \\
{{Toxicity --- a property of a communicative event in which a communicative act, interpreted by an audience within a given context, is perceived as violating relevant moral or interactional norms and thereby induces stress.}}
\end{quote}

\begin{quote}
\textbf{Definition 2 (Toxic speech).} \\
{{Toxic speech --- a linguistic communicative event in which an utterance, interpreted by an audience within a given context, is perceived as violating relevant moral or interactional norms and thereby induces stress.}}
\end{quote}

{These definitions deliberately assign toxicity to interpreted communicative events rather than to text strings in isolation. A surface text} \(x\) {is therefore not toxic per se; it could be a part of a toxic communicative event in relation to an audience, a context of interpretation, and the norms through which it is evaluated. This is why we distinguish between \emph{toxicity} and \emph{toxic speech}. Toxicity is the broader property of communicative events and may, in principle, apply to many kinds of communicative acts: sharing an image, posting a video, using a gesture, displaying a symbol, or combining several modalities at once. Toxic speech is the linguistic realisation of the concept.}

\section{Formalisation of the framework}
\label{sec:ch8_objects}

We now formalise the framework more explicitly.

Let a \textbf{unit communicative event} for audience unit \(a\) be denoted by
\[
e_a = (x, C, a),
\]
where:
\begin{itemize}
    \item \(x\) is the communicative act itself (e.g. an utterance, message, post, reply, image, gesture, or other signal);
    \item \(C\) is the context in which \(x\) is interpreted;
    \item \(a\) {is the relevant audience unit. An audience unit} \(a\) {may be an individual reader, a subgroup, or a community, depending on the level of analysis.}
\end{itemize}


We define \textbf{context} as a structured collection of conditions relevant to interpretation:
\[
C = \{C^{(k)}\}_{k=1}^{m}
\]
including, but not limited to:
\begin{itemize}
    \item $C^{v}$ -- \textbf{verbal context}: surrounding discourse, co-text, thread history, quoted material, and other linguistic material that constrains interpretation;
    \item $C^{s}$ -- \textbf{situational context}: speaker, target, platform, institutional setting, timing, role relations, and other circumstances of use;
    \item $C^{g}$ -- \textbf{cognitive context}: inferred intentions, expectations, shared background knowledge, emotional stance, and perspective of participants;
    \item $C^{c}$ -- \textbf{cultural context}: norms, conventions, taboos, community-specific meanings, and historically situated expectations;
    \item $C^{r}$ -- \textbf{relational context}: prior relationship between speaker, target, and audience, including familiarity, hierarchy, dependency, intimacy, and prior conflict.
\end{itemize}

\subsection{Perceived norm violation}
The framework now requires two components: perceived norm violation, and stress response.
Let
\[
\nu(e_a) \in [0,1]
\]
denote the degree to which audience unit \(a\) perceives \(x\) in context \(C\) as violating relevant norms.

A value of $\nu(e_a)=0$ means that the event is perceived as norm-compatible; a value near $1$ means that it is perceived as strongly norm-violating. This quantity is intentionally defined as \emph{perceived} violation, not objective violation. CSF models social interpretation, not metaphysical moral truth.

The specification of the relevant norms is therefore a modelling choice. Existing systems already make such choices, albeit implicitly: through dataset selection, annotation guidelines, annotator demographics, platform policy and other design decisions. CSF requires them to be made explicit and declared as part of the measurement design.

This explicit specification makes it possible to use existing models to operationalise this component of the framework: text-intrinsic toxicity detectors can be understood as estimating perceived norm violation under an assumed or averaged context. Provided that this assumed context is made explicit and is applicable to the communicative event being analysed, such models may serve as operationalisations of \(\nu(e_a)\). CSF therefore does not discard existing toxicity detectors; rather, it states that they capture only one component of toxicity.

\subsection{Stress response}

Let now
\[
\sigma(e_a) \in [0,1]
\]
denote the degree of stress in audience unit \(a\) that is attributable to the interpretation of
\(x\) in context \(C\).

Unlike broad labels such as "offensive" or "rude", stress has a comparatively well-characterised biological basis and a long measurement tradition. Across psychology and neuroscience, stress is understood as a family of physiological and behavioural responses arising when an individual appraises a situation as threatening, boundary-violating, or socially unsafe \citep{mcewen1998,thayer2012}. 

Stress can be estimated through multiple classes of indicators, each involving a different trade-off between fidelity and scalability. All of these signals can be both measured and predicted: 

\textbf{Direct physiological measures}, such as levels of cortisol, adrenaline or noradrenaline are biologically faithful but require invasive sampling and are impractical outside controlled settings. 

\textbf{Non-invasive biosignals} - including EEG activity, heart-rate variability, galvanic skin response, respiration, pupil dilation, and facial or vocal affect - provide more scalable approximations and have been shown to strongly correlate with stress and social-threat responses \citep{sharma2012objective,soc_stress_biosignals}. Such measurements are feasible in focus-group studies and controlled experiments. Some signals can also be collected using consumer wearable devices, such as smartwatches, making them increasingly practical in applied settings, including health-support systems.

\textbf{Behavioural markers of stress} provide the most scalable class of indicators. Stress often surfaces in interaction through reaction patterns, shifts in tone, escalation, withdrawal, or affective language in replies. Prior work suggests that collective response dynamics can serve as indirect indicators of social stress and interactional tension in online discussion \citep{Amir,Aleksandric_2024}. Behavioural proxies do not measure internal physiology directly, but they capture how social tension becomes publicly legible in interaction - precisely the level at which toxicity becomes consequential for communities and moderation decisions.

\subsection{Individual toxicity contribution}

We now define the \textbf{toxicity contribution} of unit event \(e_a\) as
\[
\tau(e_a) = g\big(\nu(e_a), \sigma(e_a)\big).
\]
\[
g : [0,1]^2 \rightarrow [0,1]
\]
where \(g\) is an interaction function combining perceived norm violation and stress response.

{At the level of the framework,} \(g\) {is deliberately left open. CSF is intended to specify the conceptual relation between perceived norm violation and induced stress, not to prescribe a single measurement instrument. However, admissible realisations of} \(g\) {should preserve the core assumption of the framework:} \(g\) {should be monotone non-decreasing in both arguments and should assign zero toxicity when either component is absent.
That is, for all} \(v,s\in[0,1]\):
\[
g(0,s)=0,\qquad g(v,0)=0,
\]

{Different downstream implementations may therefore instantiate} \(g\) {depending on the measurement setting, available signals, and moderation policy.}

{Examples include the product} \(g(v,s)=vs\), {which treats toxicity as jointly dependent on both
components, or} \(g(v,s)=\min(v,s)\), {which makes toxicity limited by the weaker component.}

The framework assumes that toxicity arises when both components are present, even though particular operationalisations may differ in how they realise this interaction. This separates toxicity from adjacent cases such as:
\begin{itemize}
    \item Norm violation without stress;
    \item Stress without a perceived norm violation, as in many forms of necessary but non-toxic discomfort.
\end{itemize}

\subsection{Group-level toxicity}
\label{sec:ch8_event_toxicity}

To move from unit to group-level toxicity, {we must aggregate toxicity contributions across the relevant audience units} \(
A=\{a_1,\ldots,a_n\}
\). {Different aggregation operators correspond to different assumptions about how experiences should count at the group level.}

{As one possible aggregation rule, we may consider a weighted mean. Let} $w_a \geq 0$ be an audience weight {assigned to audience unit} \(a\), {with} \(\sum_{a \in A} w_a = 1\). Depending on the measurement setting, weights may encode exposure, relevance to the communicative event, or the sampling distribution over audience units. In protective measurement settings, weights may also be adjusted to give greater importance to vulnerable or directly targeted audience units.

Then the group-level toxicity of communicative event $e$ is

\[
T(x,C,A)
=
\sum_{a \in A} w_a \tau(e_a)
\]

Where, for each audience unit \(a \in A\), \(e_a=(x,C,a)\).

CSF conceptually separates the unit-level toxicity contribution \(\tau(e_a)\) from the group-level toxicity score \(T(x,C,A)\).

{In open online settings, the relevant audience may not be fully known at the time of prediction. In such cases, the audience set} \(A\) {should be understood as a modelling choice: it may represent expected viewers, the affected community, vulnerable subgroups, target-group members, or another policy-relevant audience. The framework does not require the actual audience to be known in advance, but it does require the system to state whose interpretation is being estimated.}

\section{Implications for Modelling and Evaluation}

CSF implies a change of view on how toxicity detection systems should be tested. They should be tested on whether their predictions change appropriately not only when the communicative act is changed, but also the context \(C\), the relevant audience unit \(a\), or the available evidence about the stress response changes. This requires testing both \emph{sensitivity} to changes that alter the interpretation of an utterance and \emph{invariance} to changes that should not affect it.

We propose five contrastive evaluation slices:

\begin{itemize}
    \item {\textbf{Same act, different context.}
    This slice holds the surface utterance fixed while varying audience, relation, institutional setting, cultural background, or prior interaction. It tests whether a system recognises that the same words can function as affiliation, humour, harassment, quotation, or humiliation depending on context.}

    \item {\textbf{Different form, same communicative harm.}
    This slice varies the surface form while preserving the interpreted communicative act. Explicit abuse, coded abuse, adversarially encoded insults, euphemisms, dog whistles, indirect threats, and pragmatic implications may produce similar perceived norm violations and stress responses. A system should therefore not rely only on overt toxic markers.}

    \item {\textbf{Missing context and partial observability.}
    This slice removes or withholds information about the speaker, audience, interaction history, relevant norms, or stress response. It tests whether systems represent uncertainty, abstain, or defer to human review when evidence is insufficient, rather than producing overconfident estimates from isolated text alone. Missing evidence should not be treated as evidence that either component is absent.}

    \item {\textbf{Use of stress evidence.}
    This slice varies the availability, reliability, and consistency of evidence about stress. It tests whether systems incorporate such evidence appropriately while distinguishing stress-related signals from controversy, majority disapproval, coordinated reporting, or other reactions that do not by themselves establish toxicity. Systems should also represent uncertainty when available indicators are noisy or conflicting.}

    \item {\textbf{Measurement-policy separation.}
    This system-level slice holds the estimated perceived norm violation and stress response fixed while varying platform rules, legal regimes, or institutional settings. It tests whether systems distinguish measurement outputs from enforcement decisions such as warning, downranking, removal, or escalation. The same measured communicative event may legitimately produce different actions under different policies without changing the underlying CSF estimate.}
\end{itemize}
\subsection{Discussion}
\label{sec:csf_desiderata}

CSF is intended as a measurement framework for toxicity, not as an all-purpose moral theory or enforcement policy. It estimates situated communicative harm, but does not by itself determine what a platform, institution, or legal system should do with that estimate. Within this scope, CSF improves the measurability, contextual validity, calibration, and auditability of toxicity detection.

First, CSF improves \textbf{measurability} by specifying two components that must be assessed: perceived norm violation and attributable stress. Stress may be grounded in physiological, behavioural, or interactional indicators, while perceived norm violation may be grounded in identifiable standards such as platform rules, community norms, legal categories, conversational expectations, or explicit annotation guidelines. CSF therefore requires evidence for both components rather than inferring toxicity from surface text alone.

Second, CSF is explicitly \textbf{context-aware}. Toxicity is represented not as a property of text alone, but as a property of \(e_a=(x,C,a)\), where the communicative act, context, and audience unit jointly define the object being evaluated. Context is therefore constitutive of the target rather than an optional feature added to a text classifier.

Third, CSF supports context- and policy-dependent \textbf{calibration}. A single universal threshold or interaction function (g) is unlikely to be appropriate across all communities, audiences, and enforcement settings. Instead, an implementation should specify the communicative act, context, audience, aggregation rule, and policy setting to which its toxicity estimate applies.

Finally, CSF improves \textbf{auditability} by making the components and assumptions of the measurement explicit. A CSF-based system should report the evidence used for perceived norm violation and stress, whose perspective is represented, how audience-level contributions are aggregated, and where uncertainty remains. This does not make toxicity measurement simple, but it makes its evidential basis and design choices visible.

\section{Conclusion}
\label{sec:conclusion}

Toxicity should not be conceptualised as a stable latent property to be read off from a given text. Harmful communication is socially interpreted, relational, and context-sensitive. Any approach that reduces it to decontextualised lexical content risks producing both false negatives and false positives.

This paper introduced the Contextual Stress Framework (CSF) in response to this construct problem. Under CSF, toxicity is treated as a property of \emph{a communicative event}, comprising the communicative act, its context, and the relevant audience. This reframing does not deny the usefulness of text-based signals, but argues that they do not capture the phenomenon fully. Contextual variation should therefore not be treated as noise around a stable label, but as part of the construct itself.

Toxicity detection should consequently be approached as the design of robust, context-aware, and governable safety systems for human communication. If such systems are to function as infrastructure, they must be evaluated not only by how well they reproduce static labels, but also by how well they remain aligned with communicative harm under realistic variation, adversarial pressure, and shifting social contexts.

\section{Limitations}
\label{sec:limitations}

CSF is a measurement framework rather than a complete operational system. It specifies the object of measurement and the relation between perceived norm violation and induced stress, but it does not prescribe a universal instrument for estimating either component, a single interaction function \(g\), or a single aggregation rule. Each implementation must therefore justify its chosen indicators, calibration procedure, and intended population.

Although stress is objectively measurable, its measurement remains instrument-dependent. Physiological, behavioural, and interactional indicators differ in fidelity, availability, temporal resolution, and susceptibility to confounding factors. In naturalistic settings, a further challenge is attribution: an observed stress response may coincide with a communicative event without being caused by it. Operationalisations of \(\sigma(e_a)\) must therefore establish, as far as the setting permits, that the measured response is attributable to the interpretation of \(x\) in context \(C\).

CSF also depends on information that may be only partially observable. The relevant audience, prior interaction, social relationship, applicable norms, and stress response may be unavailable at inference time. This does not invalidate the framework, but it limits the confidence with which a communicative event can be assessed. Missing evidence must be represented as uncertainty rather than interpreted as evidence that perceived norm violation or stress is absent.

Finally, A CSF estimate does not by itself determine whether content should be permitted, warned, downranked, removed, or escalated. Such decisions additionally depend on institutional policy, legal constraints, proportionality, and available remedies. The framework is therefore intended to inform accountable decision-making, not to replace it.

\section{Ethical Considerations}
\label{sec:ethics}

Contextual modelling creates ethical risks in addition to potential benefits.
Estimating audience characteristics, relationships, community membership,
emotional reactions, or behavioural responses may require sensitive personal
data. Systems implementing CSF should follow data-minimisation principles and
avoid inferring protected attributes or psychological states unless such
inferences are necessary, justified, and appropriately governed.

Particular
care is required when modelling reclaimed language, dialectal variation,
counterspeech, activism, whistleblowing, satire, and speech produced by
minority communities.

The separation between measurement and enforcement is therefore ethically
important. An estimate of contextual harm should not automatically determine
removal or punishment. High-impact actions should incorporate applicable policy,
the rights and vulnerability of affected participants, the reliability of the
available context, and opportunities for human review or appeal.

CSF may also be misused to justify intrusive surveillance or automated inference
of individual stress. The framework is intended to support accountable
measurement of communicative events, not continuous monitoring of users. Implementations should report which signals are
collected, whose perspective they represent, how uncertainty and disagreement
are handled, and how measurements influence policy decisions.

\begingroup
\sloppy
\bibliography{custom}
\endgroup

\appendix

\section{Definitions across domains}
\label{appA}

We examine how hate speech and related forms of harmful language are defined across legal, industry, and academic contexts. In doing so, we compare major definitional traditions, synthesise typological frameworks that account for definitional pluralism, and identify recurring limitations associated with different definitional families.

\subsection{Legal and human-rights frameworks}

Legal systems rarely offer a single, unified definition of hate or toxic speech. Instead, they regulate partially overlapping categories such as incitement to hatred, discriminatory harassment, group defamation, and threats, typically anchored in protected characteristics and shaped by local constitutional traditions.

At the European level, the Council of Europe Recommendation No.\ R (97) 20 defines hate speech as forms of expression that "spread, incite, promote or justify hatred based on intolerance" \citep{coe1997r9720}. More recently, Recommendation CM/Rec(2022)16 adopts a broader framing and explicitly distinguishes layers of severity, with guidance on proportional responses \citep{coe2022cmrec16}.

Similarly, the United Nations Strategy and Plan of Action on Hate Speech adopts a working definition encompassing communication that "attacks or uses pejorative language towards a person or group on the basis of identity factors", while emphasising that there is no universally agreed international legal definition \citep{un2019strategy_hate_speech}. Binding instruments such as ICERD require states to prohibit certain forms of "racist propaganda and incitement", while leaving substantial room for domestic constitutional balancing \citep{icerd1965}.

Importantly, contemporary human-rights practice often conceptualises hate speech along a spectrum rather than as a binary category. The UN Rabat Plan of Action proposes a structured threshold test for incitement assessments (context, speaker, intent, content, extent, and likelihood), encouraging severity-sensitive responses rather than one-size-fits-all criminalisation \citep{rabat2012}.

\subsection{Platform governance}

Platform definitions emerge from a distinct optimisation problem: enforcing rules over massive volumes of content under constraints of scale, speed, and reputational risk. Accordingly, platform policies tend to be enumerative and operational: they specify protected classes, define prohibited "attack" types, and carve out procedural exceptions (e.g., counterspeech, quotation, satire) to support consistent enforcement.

From a research perspective, these policies are less a theory of hate speech than an operational enforcement instrument: they encode thresholds, exceptions, and institutional risk trade-offs, often diverging from both legal doctrine and dataset label schemes \citep{kiritchenko2021ethical_human_rights}.

For example, Meta defines "hateful conduct" as direct attacks against people (rather than ideas or institutions) on the basis of protected characteristics, and operationalises this definition through categories such as dehumanising speech, harmful stereotypes, statements of inferiority, and calls for exclusion \citep{meta_hateful_conduct}. X (formerly Twitter) prohibits hateful conduct targeting people based on protected categories, including slurs and tropes intended to degrade or reinforce harmful stereotypes; enforcement may vary with severity and context \citep{x_hateful_conduct_policy}. LinkedIn similarly prohibits hateful or derogatory content that attacks, intimidates, dehumanises, or incites hatred or violence against protected groups \citep{linkedin_hateful_derogatory,linkedin_professional_policies}.

\subsection{Academic and computational practice}

In computational research, definitions most often take the form of annotation guidelines, with the goal not to settle philosophical disputes, but to generate labels that are sufficiently consistent to train and evaluate models \citep{fortuna2020toxic_hateful_offensive}.

A prominent example is the family of Wikipedia-based toxicity datasets and the Perspective API framing, which operationalise toxicity in terms of comments that are rude, disrespectful, or unreasonable and that risk driving participants away \citep{def_old,perspectiveapi}. Other datasets adopt narrower identity-based formulations; for instance, \citet{davidson2017hate_offensive} separates hate speech from offensive language and shows that lexical cues alone often collapse that distinction.

Shared-task traditions further illustrate how definitions are "built" by layering criteria (e.g., offensiveness, targeting, protected group status), as in OffensEval \citep{zampieri2019offenseval}. While such schemes can be internally coherent, they often lack cross-contextual stability: datasets encode particular community norms, dialects, and historical moments, and the field frequently conflates adjacent phenomena (hate speech, harassment, profanity, incivility), limiting comparability and generalisation \citep{fortuna2020toxic_hateful_offensive,vidgen2021garbage,mnassri2024multilingual_offensive_survey}.

\end{document}